\documentclass[runningheads]{llncs}
\usepackage{amsmath}
\usepackage[T1]{fontenc}
\usepackage{hyperref}
\usepackage{graphicx}
\usepackage{latexsym}
\usepackage{mathabx}
\usepackage{enumitem}
\usepackage{comment}
\usepackage{color}
\usepackage{float}
\usepackage{ragged2e}
\usepackage{subcaption} 
\usepackage{booktabs,makecell,multirow,tabularx}

\usepackage{caption}
\usepackage{xcolor}
\begin{document}
\title{AMPLE: Emotion-Aware Multimodal Fusion Prompt Learning for Fake News Detection}

\titlerunning{AMPLE: Emotion-Aware Multimodal Fusion Prompt Learning}


%

\author{Xiaoman Xu \and
Xiangrun Li \and
Taihang Wang 
\and Ye Jiang\thanks{Corresponding author: ye.jiang@qust.edu.cn}
}
\authorrunning{X. Xu et al.}
 \institute{College of Information Science and Technology, \\ Qingdao University of Science and Technology, \\ China 
 }
%
\maketitle              
\begin{abstract}

Detecting fake news in large datasets is challenging due to its diversity and complexity, with traditional approaches often focusing on textual features while underutilizing semantic and emotional elements. Current methods also rely heavily on large annotated datasets, limiting their effectiveness in more nuanced analysis. To address these challenges, this paper introduces Emotion-\textbf{A}ware \textbf{M}ultimodal Fusion \textbf{P}rompt \textbf{L}\textbf{E}arning (\textbf{AMPLE}) framework to address the above issue by combining text sentiment analysis with multimodal data and hybrid prompt templates. This framework extracts emotional elements from texts by leveraging sentiment analysis tools. It then employs Multi-Head Cross-Attention (MCA) mechanisms and similarity-aware fusion methods to integrate multimodal data. The proposed AMPLE framework demonstrates strong performance on two public datasets in both few-shot and data-rich settings, with results indicating the potential of emotional aspects in fake news detection. Furthermore, the study explores the impact of integrating large language models with this method for text sentiment extraction, revealing substantial room for further improvement. The code can be found at :  \url{https://github.com/xxm1215/MMM2025_few-shot/}.

\end{abstract}
\section{Introduction}
\vspace{-0.2cm}
Information spreads rapidly through social media platforms such as Twitter, Facebook, and Weibo \cite{jiang2023similarity}. However, the proliferation of digital news on these platforms presents significant challenges to the effective regulation of fake news, which undermines the credibility of traditional media and increases associated economic and political risks \cite{richler2023social}. Traditional manual review methods \cite{castillo2011information,yang2012automatic} are increasingly inadequate in handling the massive flow of information, making it particularly difficult to detect emerging themes in the news.

\begin{figure}[H]
\centering

\includegraphics[width=8cm]{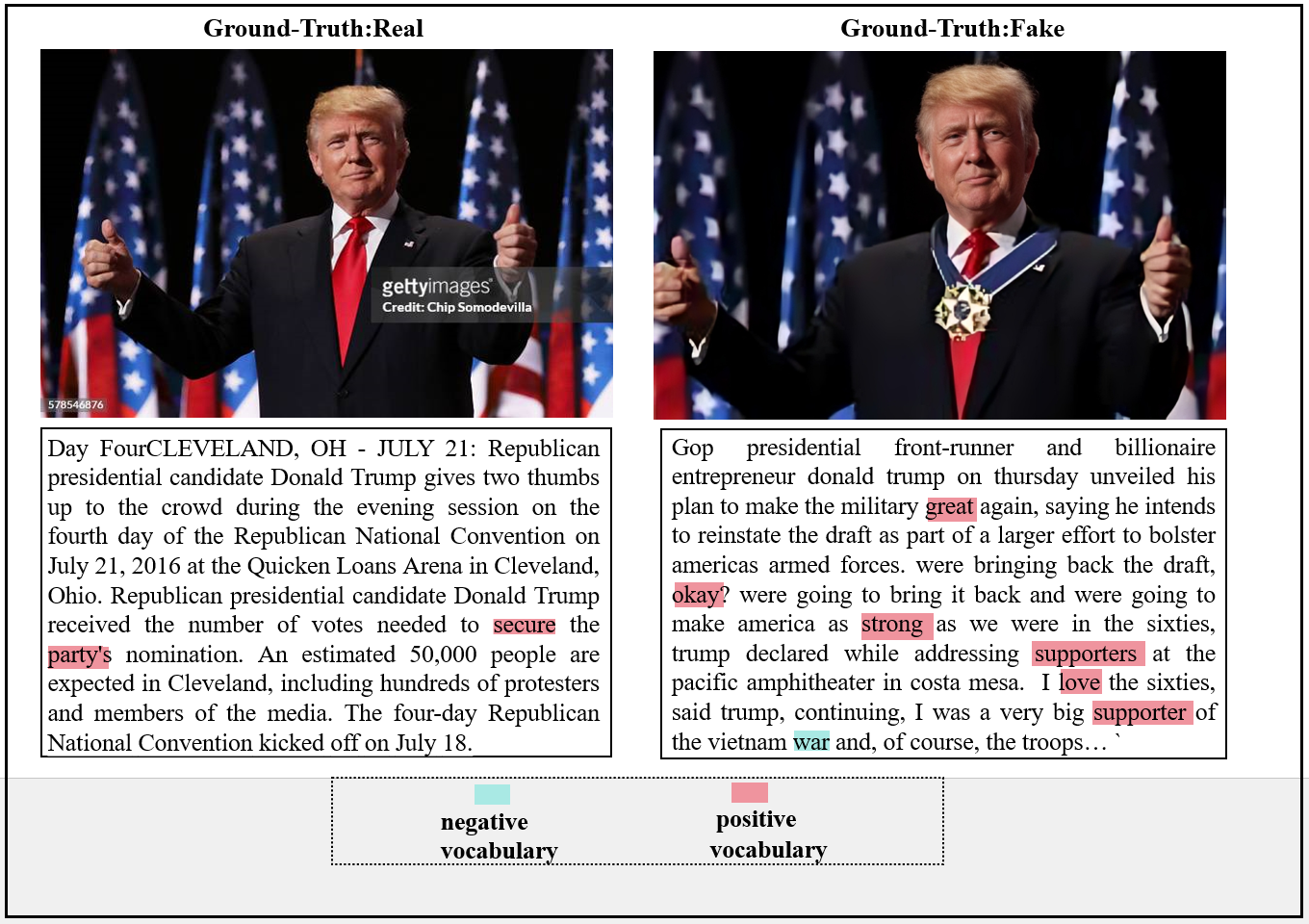}
\caption{This example illustrates the differences in emotion between real and fake news, with various colors representing distinct emotional states. The example is annotated with positive and negative sentiments based on a lexicon using the LIWC tool.
}
\label{fig:1}
\end{figure}

In response to these challenges, recent Fake News Detection (FND) models have started incorporating semantic features alongside traditional lexical and stylistic ones \cite{liu2024emotion,zhou2023does}.
Notably, the overall emotion, negativity, and expressions of anger are higher in fake news compared to real news \cite{zhou2023does}. As illustrated in Figure \ref{fig:1}, fake news often employs a greater number of emotionally charged words, thereby influencing the overall emotion distribution. Presently, research \cite{li2023multi,luvembe2023dual} on FND based on sentiment analysis focuses on exploring the relationship between sentiment and news content, using Deep Learning (DL) techniques or neural network models to extract emotional elements from text to enhance multimodal FND. However, these studies \cite{giachanou2021impact,iwendi2022covid} depend heavily on large amounts of labeled data, and constructing such datasets is highly labor- and resource-intensive. Additionally, common methods \cite{han2022ptr,jiang2023similarity,schick2020exploiting} in multimodal FND, such as cascading, addition, and specially designed neural networks, primarily facilitate the fusion of surface features but may lose deeper multimodal information.
\vspace{-0.8cm}
\vspace{-0.8cm}

To address the aforementioned challenges, we introduce strategies such as few-shot learning \cite{snell2017prototypical}  and prompt learning \cite{schick2020exploiting} to mitigate the issue of limited labeled data. We propose a novel framework for FND called the Emotion-\textbf{A}ware \textbf{M}ultimodal Fusion \textbf{P}rompt \textbf{L}\textbf{E}arning (\textbf{AMPLE}). The AMPLE framework incorporates a hybrid prompting template that combines fixed textual prompts with tunable vectors, specifically designed to meet the requirements of FND. The framework employs a pre-trained CLIP-ViT-B/32\footnote{\url{https://github.com/openai/CLIP}} (CLIP) model to extract features from both text and images, while a Sentiment Analysis Tool (SAT) is used to extract emotional elements from text with pronounced emotional tendencies or subjectivity, which are then integrated into the multimodal features. Additionally, we utilize the Multi-head Cross-Attention (MCA) mechanism. This mechanism calculates and normalizes the semantic similarity between features to fine-tune the strength of the fused multimodal representations.

We conduct evaluations on two publicly available datasets, PolitiFact and GossipCop \cite{jiang2023team,shu2020fakenewsnet}, demonstrating the benefits of AMPLE in both few-shot and data-rich settings. AMPLE generally outperforms existing methods in terms of macro F1 (F1) and accuracy (Acc). Moreover, previous studies \cite{feng2023affect,zhang2023enhancing} have explored improving sentiment analysis accuracy using Large Language Models (LLMs). Building on these findings, we introduce LLM to extract emotional elements and integrate them with our approach to further enhance the effectiveness of multimodal FND.

\begin{itemize}
    \item This paper introduces the AMPLE framework, integrating sentiment analysis into multimodal FND, enhancing the ability to verify news authenticity by analyzing emotional tendencies.
    \item This paper explores through experimentation the impact of emotional factors on detecting fake news and verifies the positive correlation between emotional elements and false information.
    \item 
    This paper demonstrates that extracting sentiment information employing LLM can enhance the performance of AMPLE in FND, while revealing that LLM still has the potential for further improvement.
   
\end{itemize}

\section{Related work}
\vspace{-0.2cm}
\subsection{Sentiment-based FND}
\vspace{-0.2cm}

Sentiment analysis has been explored across various domains. \cite{zhou2023does} employs TextBlob\footnote{\url{https://textblob.readthedocs.io/en/dev/}} for emotion polarity and LIWC\footnote{\url{https://www.liwc.app/}} for multi-dimensional sentiment analysis, confirming that fake news exhibits higher levels of overall and negative emotion compared to real news. Research \cite{martel2020reliance} shows that emotions significantly affect judgments of fake news accuracy, with emotion-based traits enhancing rumour recognition more effectively than personality-based traits \cite{fersini2020profiling}. Negative perceptions of the political system can also impact news credibility evaluations \cite{rijo2023s}. Consequently, many studies \cite{liu2024emotion,luvembe2023dual} combine ML, DL, and advanced fusion techniques to extract or prioritize emotional elements in FND. However, these methods \cite{giachanou2021impact,iwendi2022covid} often require extensive labeled datasets, which are costly and prone to bias. Moreover, the complexity of model design and training poses additional challenges. In contrast, this paper employs SATs, which utilize sentiment lexicons and rules to compute emotional elements. SAT's straightforward API obviates the need for complex model design and training.
\vspace{-0.3cm}
\subsection{Prompt Learning FND}
\vspace{-0.2cm}

Prompt Learning leverages pre-trained models' factual and commonsense knowledge through specific templates. It can be categorized into two types: (1) \textbf{pseudo-prompted fine-tuning of masked Language Models (LMs)} and (2) \textbf{prom-pted fine-tuning of autoregressive LMs}. The first type, pseudo-prompts, introduces specific vectors into the model's input to guide outputs. For instance, \cite{lee2021towards} uses a masked LM to generate pseudo-prompts for few-shot fact-checking. \cite{ma2021template} adapts BERT's pre-training task for entity language modeling with template-free prompt adaptation. The second approach involves creating and integrating prompt templates based on task requirements. For example, \cite{li2021prefix} proposes Prefix-Tuning with continuous prompt templates for feature representation adjustment, while \cite{li2021sentiprompt} introduces SentiPrompt, a sentiment-based prompt fine-tuning method using BART. The success of prompt learning in few-shot scenarios \cite{jiang2024cross} motivates our study, leading us to integrate these techniques into our AMPLE framework to enhance detection in resource-limited settings.

\section{Methodology}
\begin{figure*}
\centering
\includegraphics[width=12cm]{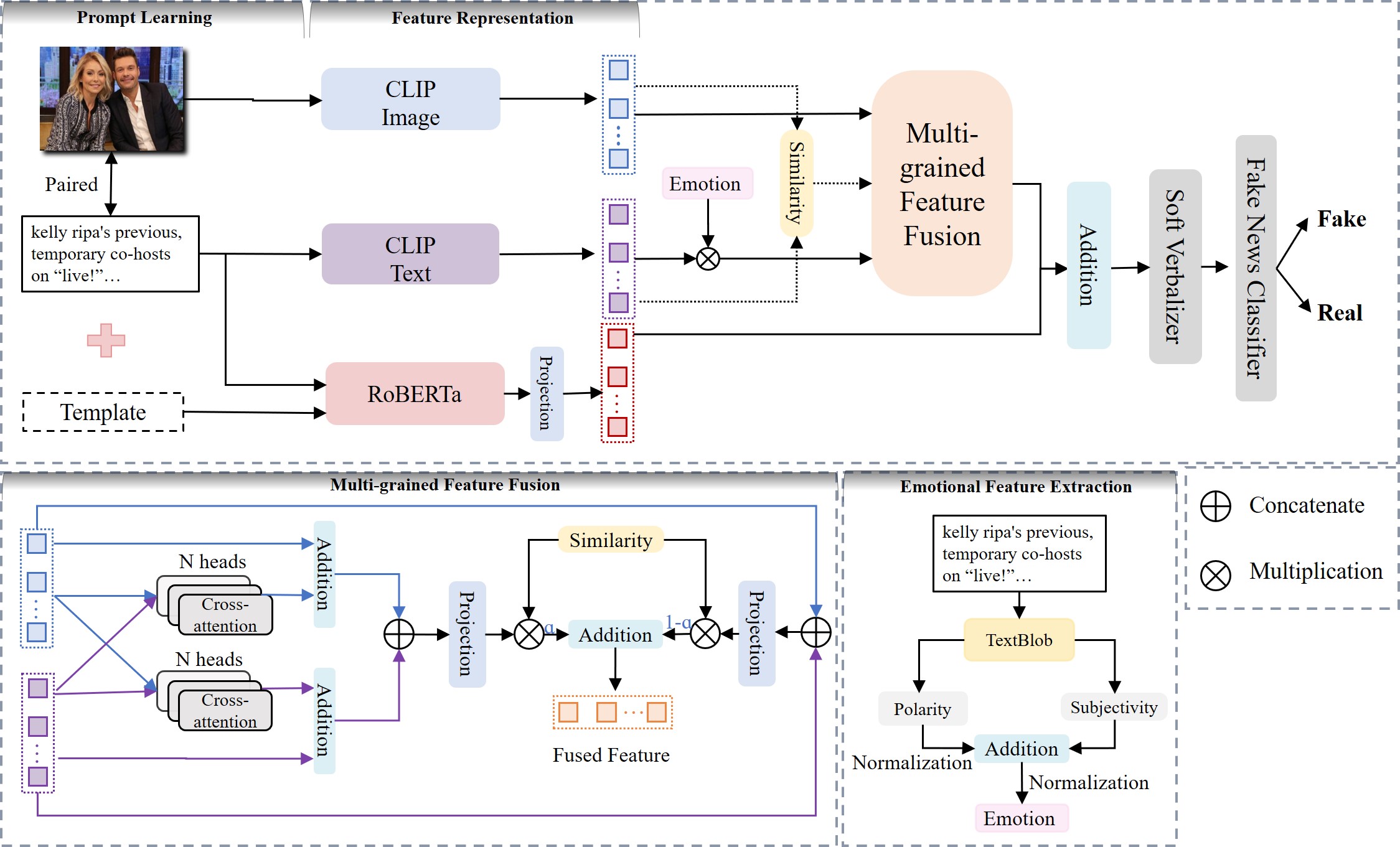}
\caption{
The overall architecture of the proposed AMPLE.
}
\label{fig:2}
\end{figure*}

\vspace{-0.2cm}
We introduce the AMPLE framework to comprehensively utilize text and image data from news articles for assessing their veracity, especially in few-shot and data-rich scenarios. The framework, outlined in Figure \ref{fig:2}, comprises feature representation, emotional elements extraction, multi-grained feature fusion, prompt learning, and task learning.
\vspace{-0.4cm}
\vspace{-0.4cm}
\subsection{Feature Representation}
We employ the CLIP model, designed for textual and visual inputs, promoting joint feature learning in a unified embedding space.  For an article with $n$ words, represented as $W=(w_1, w_2, \ldots, w_n)$, and $m$ images, denoted as $I=(i_1, i_2, \ldots, i_m)$. $T^w$ represents text features, and $V^i$ represents image features.
\vspace{-0.4cm}
 \subsection{Emotional Elements Extraction}
\vspace{-0.2cm}
To integrate sentiment information, we use SAT to extract emotional elements from text. For each news sample \( W \), we extract emotion polarity (\( p \)) and subjectivity (\( s \)), with \( p \) in \([-1, 1]\) and \( s \) in \([0, 1]\). We normalize \( p \) to \( p^* \) as shown in Eq.~\eqref{eq:1}, and combine \( p^* \) and \( s \) into a composite emotion element \( e \). To highlight emotion intensity and subjectivity, we adjust \( e \) to \( 1 \times 10^{-2} \) if \( e = 0 \). We then fuse features by multiplying the normalized \( e \) with \( T^w \) to obtain the sentiment-enhanced text embedding \( T^e \) as shown in Eq.~\eqref{eq:2}.

\begin{eqnarray}\label{eq:1}
p^*&=& \frac{p-p_\text{min}}{p_\text{max}-p_\text{min}}\\
T^e&=& (\frac{e-e_\text{min}}{e_\text{max}-e_\text{min}})T^w
\label{eq:2}
\end{eqnarray}
\vspace{-0.4cm}
\subsection{Multi-grained Feature Fusion}
\vspace{-0.2cm}

We use the MCA mechanism to explore the deep fusion between modalities \( T^e \) and \( V^i \), obtaining new representations \( V^a \) (Eq.~\eqref{eq:4}) and \( T^a \) (Eq.~\eqref{eq:5}) through the MCA mechanism:
\begin{align}
\label{3}
head_i =Att(Q_i, K_i, V_i)= softmax\left(\frac{Q_i K_i^T}{\sqrt{d_k}}\right) V_i\\
\label{eq:4}
V^a =Multi(V^i, T^e, T^e) = Cat[head_1, \ldots, head_m]\\
\label{eq:5}
T^a = Multi(T^e, V^i, V^i)= Cat[head_1, \ldots, head_m]
\end{align}

Here, the dimension \(d_k\) of the key is set to \(d/2\). Cat stands for concatenate. The decision to utilize powers of 2 (1, 2, 4, 8) as the number of attention heads \(m\) is based on the dimensionality of \(T^w\) and \(V^i\) (i.e., 512). This choice ensures that each attention head processes sub-vectors of equal dimensions, improving the efficiency of parallel computation and preventing computational waste and uneven data processing. 

We use a residual connection strategy, adding \( V^a \) and \( T^a \) to the original modal features to obtain integrated features \( V^c \) and \( T^c \). A fusion framework \( \text{mix} \) with linear transformations, batch normalization, ReLU activation, and dropout layers generates integrated feature vectors \( f_1 \) and \( f_2 \). We optimize cross-modal interactions and reduce noise by calculating and normalizing the cosine similarity (Eq.~\eqref{eq:6}) between \( T^w \) and \( V^i \), combining it with \( f_1 \) and \( f_2 \) to obtain final cross-modal features \( m_1 \) and \( m_2 \) (Eq.~\eqref{eq:7}).

\begin{eqnarray}\label{eq:6}
sim &=& \frac{T^w \cdot (V^i)^T}{||T^w|| ||V^i||}\\
\label{eq:7}
m_1, m_2 &=& \text{Sigmoid}(\text{Std}(sim)) \cdot f_1, f_2
\end{eqnarray}
\vspace{-0.4cm}
\subsection{Prompt Learning}
\vspace{-0.2cm}
In this paper, we perform pseudo-prompt tuning using the RoBERTa-base\footnote{\url{https://huggingface.co/FacebookAI/roberta-base}} model within the AMPLE framework, where the masked LM component is denoted as \textit{PLM}. In this study, we manually construct a hybrid prompt template that combines the interpretability of discrete prompts with the auto-update advantage of continuous prompts \cite{jiang2023similarity}. To simplify the process, the template \( T \) is fixed as "\textit{<head>This is a piece of <mask> news. <tail>}", which contains special "\textit{<mask>}" tokens, with "\textit{<head>}" and "\textit{<tail>}" being trainable tokens. The integrated input \( TW \) is obtained by combining the news text \( W \) with the prompt template \( T \). Subsequently, the lexical probability distribution for each token is computed using the pre-trained RoBERTa-base model and is represented by the sequence \( V \) (Eq. \eqref{8}).
\begin{eqnarray}\label{8}
V=P L M\left(TW\right)
\end{eqnarray}

\vspace{-0.4cm}
\subsection{Task Learning}
\vspace{-0.2cm}
We further optimize prompt selection in the continuous embedding space by combining \( V \) with \( m_1 \) and \( m_2 \) via residual connections. Additionally, we introduce an adjustment coefficient \( \alpha \) (where \( 0 \leq \alpha \leq 1 \)) to balance the importance of textual and visual information in the final decision-making process. The composite feature vector \( x_f \) is then computed as Eq. \eqref{9}:

\begin{eqnarray}\label{9}
x_f = \text{FC}\left(V + \alpha \cdot m_1 + (1-\alpha) \cdot m_2\right)
\end{eqnarray}
\( x_f \) is subsequently fed into a fully connected (FC) layer for further processing, ultimately determining the most suitable fill-in word for the "\textit{<mask>}" token.

The classification probability \( P(y \mid x_f) \) Eq. \eqref{10} is computed as:
\begin{equation}
P(y \mid x_f) = \frac{\exp(\theta_y^V \cdot x_f)}{\sum_{i \in C} \exp(\theta_i^V \cdot x_f)}
\label{10}
\end{equation}

where \( C \) is the set of classes, \( \theta_y^V \) is the embedding for the true label, and \( \theta_i^V \) are the embeddings for the predicted label words. Finally, the cross-entropy loss \(\theta^*\) Eq. \eqref{11} can be minimized as:
\begin{equation}
\theta^* = \arg \max_{\theta} \left(-\log P(y \mid x_f)\right)
\label{11}
\end{equation}

\section{Experiment}
\subsection{Dataset}
In our study, we utilize two publicly available FND datasets: PolitiFact (1,056 news articles) and GossipCop (22,140 news articles). We calculate the similarity between text and image using the CLIP model, retaining only the pairs with the highest similarity to ensure that each text corresponds to its related image. Referring to \cite{zhou2023does}, we employ TextBlob to evaluate the emotion polarity of news, and calculate the matching percentage of each word in the text with sentiment-related categories (such as positive, negative) through the LIWC dictionary-based method, thereby deriving overall, positive, and negative emotion scores.The dataset statistics and the mean and standard deviation of the emotion indicators are listed in Table \ref{tab:1}. The results show that fake news differs from real news in most emotion values, with real news generally showing higher emotion values. See Section \ref{4.4} for detailed analysis.


\vspace{-15pt}

\begin{table}[ht]
\centering
\caption{Preprocessed dataset statistics. The mean and standard deviation values for each emotion indicator (Overall, Positive, Negative, and Polarity emotion) are computed across all news articles within each category (mean [std]).}\label{tab:1}
\setlength{\tabcolsep}{6pt}  
\begin{tabular}{ccccc}
\toprule  
\multirow{2}{*}{\textbf{Statistics}} & \multicolumn{2}{c}{\textbf{Politifact}} & \multicolumn{2}{c}{\textbf{GossipCop}} \\
\cmidrule(lr){2-3} \cmidrule(lr){4-5}  
                                     & \textbf{fake}      & \textbf{real}      & \textbf{fake}      & \textbf{real}     \\
\midrule  
Number of news                       & 96                 & 102                & 1877               & 4928              \\
Average of words                    & 444               & 3387              & 726               & 699              \\
\midrule 
  {\textbf{Emotion Indicators}}                    & mean {[}std{]}  & mean {[}std{]}  & mean {[}std{]}  & mean {[}std{]} \\
\midrule 
Overall emotion score                      & 0.034 {[}0.029{]}  & 0.036 {[}0.015{]}  & 0.040 {[}0.023{]}  & 0.043 {[}0.025{]} \\
Positive emotion score                     & 0.020 {[}0.026{]}  & 0.022 {[}0.020{]}  & 0.027 {[}0.021{]}  & 0.031 {[}0.024{]} \\
Negative emotion score                     & 0.012 {[}0.015{]}  & 0.014 {[}0.010{]}  & 0.013 {[}0.012{]}  & 0.011 {[}0.013{]} \\
 Polarity emotion score                  & 0.091 {[}0.158{]}  & 0.100 {[}0.064{]}  & 0.485 {[}0.074{]}  & 0.500 {[}0.074{]} \\
\bottomrule  
\end{tabular}
\end{table}


\vspace{-0.5cm}
\subsection{Implementation Details}
\vspace{-0.2cm}
Our training process employs the AdamW optimizer with a cross-entropy loss function, an initial learning rate of $3 \times 10^{-5}$, and a decay rate of $1 \times 10^{-3}$ over 20 epochs. We utilize TextBlob's SAT to extract emotion elements from the text. As shown in Table \ref{tab:2}, the model's performance is evaluated under both few-shot and data-rich settings. In the few-shot setting, PolitiFact is trained on small sample sizes (n in [2, 4, 8, 16, 50]), and GossipCop on (n in [2, 4, 8, 16, 100]), with five repetitions using different seeds to ensure robustness. In the data-rich setting, the dataset is split into training, validation, and testing sets with a ratio of 8:1:1, and performance metrics are averaged after removing the highest and lowest values.

\subsection{Baseline}
\vspace{-0.2cm}
The proposed AMPLE framework is compared with nine models on the FND dataset. The specific model comparisons conducted include:
\begin{itemize}
    \item \textbf{Unimodal Approaches:} \textbf{LDA-HAN:} \cite{jiang2020comparing} integrates LDA with a Hierarchical Attention Network. \textbf{T-BERT:} \cite{bhatt2022fake} employs a cascaded triplet BERT architecture.
    \item \textbf{Multimodal Approaches:} \textbf{SAFE:} \cite{zhou2020similarity} transforms images into textual descriptions.
 \textbf{RIVF:} \cite{9661684} merges VGG and BERT with attention. \textbf{SpotFake:} \cite{singhal2019spotfake} uses VGG and BERT for feature extraction. \textbf{CAFE:} \cite{chen2022cross} employs fuzziness-aware feature aggregation.
    \item \textbf{Standard Fine-tuning Approaches:} \cite{jiang2023similarity} uses fine-tuning on RoBERTa.
    \item \textbf{Few-shot Learning Approaches:} \textbf{P\&A:} \cite{wu2023prompt} jointly leverage the pre-trained knowledge in PLMs and the social contextual topology. Due to the absence of user data in our dataset, the user engagement threshold is adjusted to 50 when training with the other party's dataset. \textbf{SAMPLE:} \cite{jiang2023similarity} integrates hybrid prompt learning templates with similarity-aware fusion.
\end{itemize}

\vspace{-0.6cm}
\subsection{Results}
\label{4.4}
\vspace{-0.2cm}

Table \ref{tab:2} presents the results of the proposed AMPLE approach compared to the baseline model. The model demonstrates superior performance on the PolitiFact dataset relative to GossipCop. Combined with the statistical analysis in Table \ref{tab:1}, we observe that a higher proportion of overall emotion and negative emotion in fake news than in real news, which contradicts some previous studies \cite{zhou2023does,martel2020reliance}, where it is found that fake news typically exhibits higher positive emotion and lower negative emotion. 

However, our study reveals that the ratio of affective polarity to positive emotion is indeed lower in fake news compared to real news, which aligns with previous findings. This emotion profile may pose challenges for models in accurately capturing and utilizing it, especially in the GossipCop dataset. The higher emotion polarity observed in the PolitiFact dataset likely enables models to extract relevant features more efficiently for training and prediction. Overall, AMPLE consistently outperforms in nearly all scenarios, particularly in terms of the F1 score, where it significantly surpasses other baseline models. This further validates the importance of emotional elements in FND and underscores the necessity for effective multimodal fusion.
\vspace{-0.2cm}
\begin{table}[H]
\caption{Overall macro-F1 (F1) and accuracy (Acc) between baseline and AMPLE. \textbf{Bolded} indicates the first-ranked data, while \underline{underlined} represents the second-ranked data.}
\label{tab:2}
\centering
\setlength{\tabcolsep}{3pt}
\begin{tabular}{llllllll}
\bottomrule
 & \multirow{2}{*}{\textbf{PolitiFact}} & \multicolumn{5}{c}{\textbf{few-shot(F1/Acc)}} & \multirow{2}{*}{\textbf{data-rich}} \\\cline{3-7}
     &       & \multicolumn{1}{c}{\textbf{2}} & \multicolumn{1}{c}{\textbf{4}} & \multicolumn{1}{c}{\textbf{8}} & \multicolumn{1}{c}{\textbf{16}} & \multicolumn{1}{c}{\textbf{50}}  \\ 

\hline

&LDA-HAN  & 0.37/0.39 & 0.42/0.43& 0.44/0.47& 0.48/0.52& 0.61/0.63& 0.70/0.74\\
 & T-BERT  & 0.43/0.50& 0.45/0.57& 0.50/0.54& 0.50/0.54& 0.69/0.69& 0.71/0.75\\
 & SAFE & 0.19/0.19& 0.21/0.21& 0.29/0.27& 0.33/0.49& 0.46/0.56& 0.64/0.65\\
 & RIVF & 0.35/0.48& 0.43/0.51& 0.42/0.48& 0.40/0.47& 0.43/0.49& 0.43/0.45\\
 & Spotfake  & 0.37/0.49 & 0.46/0.52& 0.47/0.54& 0.56/0.59& 0.73/0.73& 0.71/0.73\\
 & CAFE & 0.30/0.39& 0.37/0.47& 0.45/0.46& 0.47/0.49& 0.52/0.61& 0.67/0.67\\
 & FTRB & 0.46/0.52& 0.51/\underline{0.63}& 0.60/0.63& \underline{0.68} /\underline{0.70}& 0.77/\underline{0.81}& 0.79/\underline{0.84}\\
 & P\&A & \underline{0.49}/\underline{0.58}& \underline{0.57}/0.62& 0.61/0.63& 0.62/0.64& 0.57/0.61& 0.63/0.66\\
 & SAMPLE & 0.47/0.56& 0.56/0.61& \underline{0.62}/\underline{0.66}& 0.67/\underline{0.70}&\textbf{0.81}/\textbf{0.82}& \underline{0.80}/0.81\\
 & AMPLE & \textbf{0.66}/\textbf{0.67}& \textbf{0.65}/\textbf{0.67}& \textbf{0.70}/\textbf{0.71}& \textbf{0.77}/\textbf{0.78}& \underline{0.80}/0.80& \textbf{0.90}/\textbf{0.90}\\
\hline
&\multirow{2}{*}{\textbf{GossipCop}} & \multicolumn{5}{c}{\textbf{few-shot(F1/Acc)}} & \multirow{2}{*}{\textbf{data-rich}} \\\cline{3-7}
     &       & \multicolumn{1}{c}{\textbf{2}} & \multicolumn{1}{c}{\textbf{4}} & \multicolumn{1}{c}{\textbf{8}} & \multicolumn{1}{c}{\textbf{16}} & \multicolumn{1}{c}{\textbf{100}} \\
\hline
& LDA-HAN & 0.18/0.21& 0.20/0.25& 0.28/0.30& 0.34/0.40& 0.49/0.45& 0.54/0.60\\
 & T-BERT & 0.38/0.48& 0.38/\textbf{0.57}& 0.45/\textbf{0.66}& 0.45/\textbf{0.71}& 0.52/0.61& 0.61/\underline{0.74}\\
 & SAFE & 0.26/0.31& 0.33/0.41& 0.40/0.45& 0.41/0.45& 0.44/0.51& 0.55/0.64\\
 & RIVF & 0.24/0.29& 0.24/0.29& 0.24/0.29& 0.27/0.31& 0.29/0.31& 0.51/0.61\\
 & Spotfake & 0.23/0.28& 0.22/0.28& 0.23/0.28& 0.32/0.34& 0.48/0.49& 0.43/0.73\\
 & CAFE & 0.41/0.42& 0.42/0.52& 0.46/0.48& 0.47/0.56& 0.50/0.61& 0.59/0.72\\
 & FTRB & 0.39/0.41& 0.33/0.46& 0.44/0.60& 0.48/0.63& 0.52/0.64& 0.63/0.69\\
 & P\&A & \underline{0.44}/0.52& 0.36/0.50& 0.33/0.50& \textbf{0.61}/\underline{0.65}& \underline{0.58}/0.59& \underline{0.73}/0.71\\
 & SAMPLE & \textbf{0.47}/\underline{0.54}&\underline{ 0.47 }/\underline{0.56}& \underline{0.52 }/0.54& \underline{0.54}/0.60& 0.58/0.62& 0.64/0.73\\
 & AMPLE & \textbf{0.47}/\textbf{0.55}& \textbf{0.50}/\textbf{0.57}& \textbf{0.53}/\underline{0.64}& \underline{0.54}/\underline{0.65}& \textbf{0.60}/\textbf{0.66}& \textbf{0.80}/\textbf{0.85}\\
\bottomrule
\end{tabular}
\end{table}

\section{Analysis}
\vspace{-0.2cm}
\subsection{Emotional Element Calculation Study}
\vspace{-0.2cm}

We utilize three methods to explore the impact of different emotional element calculations on model performance. As shown in Table \ref{tab:4}, in the few-shot setting of the PolitiFact dataset, the \textit{AMPLE\textsubscript{p+s}} configuration achieves the best classification performance. This is likely because it combines emotional polarity with subjectivity, providing a richer understanding of the text context. Fake news often manipulates emotion and presents subjective opinions as facts, and the \textit{AMPLE\textsubscript{p+s}} configuration is better at detecting these nuances, thereby improving performance.
\begin{table}[H] 
\caption{Analysis of emotional elements in FND.
\textit{AMPLE\textsubscript{p}} represents the normalized score for calculating text emotion polarity. \textit{AMPLE\textsubscript{s}} denotes the score for evaluating text subjectivity. \textit{AMPLE\textsubscript{p+s}} integrates these approaches by adding emotion polarity to subjectivity, followed by normalization.
}\label{tab:4}
\centering
\setlength{\tabcolsep}{7.4pt} 
\renewcommand{\arraystretch}{1.5} 

\begin{tabular}{llllll}
\bottomrule

\multirow{2}{*}{\textbf{PolitiFact}} & \multicolumn{5}{c}{\textbf{few-shot (F1/Acc)} } \\\cline{2-6}
    & \multicolumn{1}{c}{\textbf{2}} & \multicolumn{1}{c}{\textbf{4}} & \multicolumn{1}{c}{\textbf{8}} & \multicolumn{1}{c}{\textbf{16}} & \multicolumn{1}{c}{\textbf{50}}  \\ 
\hline
AMPLE\textsubscript{p} & \underline{0.60}/0.60 & \underline{0.59}/\underline{0.63} & 0.68/0.68 & \underline{0.74}/\underline{0.74} & 0.76/0.77 \\  
AMPLE\textsubscript{s} & 0.59/\underline{0.62} & \underline{0.59}/0.62 & \underline{0.69}/\underline{0.70} & \underline{0.74}/\underline{0.74} & \underline{0.78}/\underline{0.78} \\  
AMPLE\textsubscript{p+s} & \textbf{0.66}/\textbf{0.67} & \textbf{0.65}/\textbf{0.67} & \textbf{0.70}/\textbf{0.71} & \textbf{0.77}/\textbf{0.78} & \textbf{0.80}/\textbf{0.80} \\
\bottomrule
\end{tabular}

\end{table}

\vspace{-0.2cm}
\subsection{Multi-grained Feature Fusion study}
\vspace{-0.2cm}

We design two strategies to evaluate the impact of MCA-computed and non-MCA-computed features on classification accuracy in AMPLE.

\begin{figure}[htbp]
    \centering
    \begin{subfigure}[b]{0.48\textwidth}
        \centering
        \includegraphics[width=\textwidth]{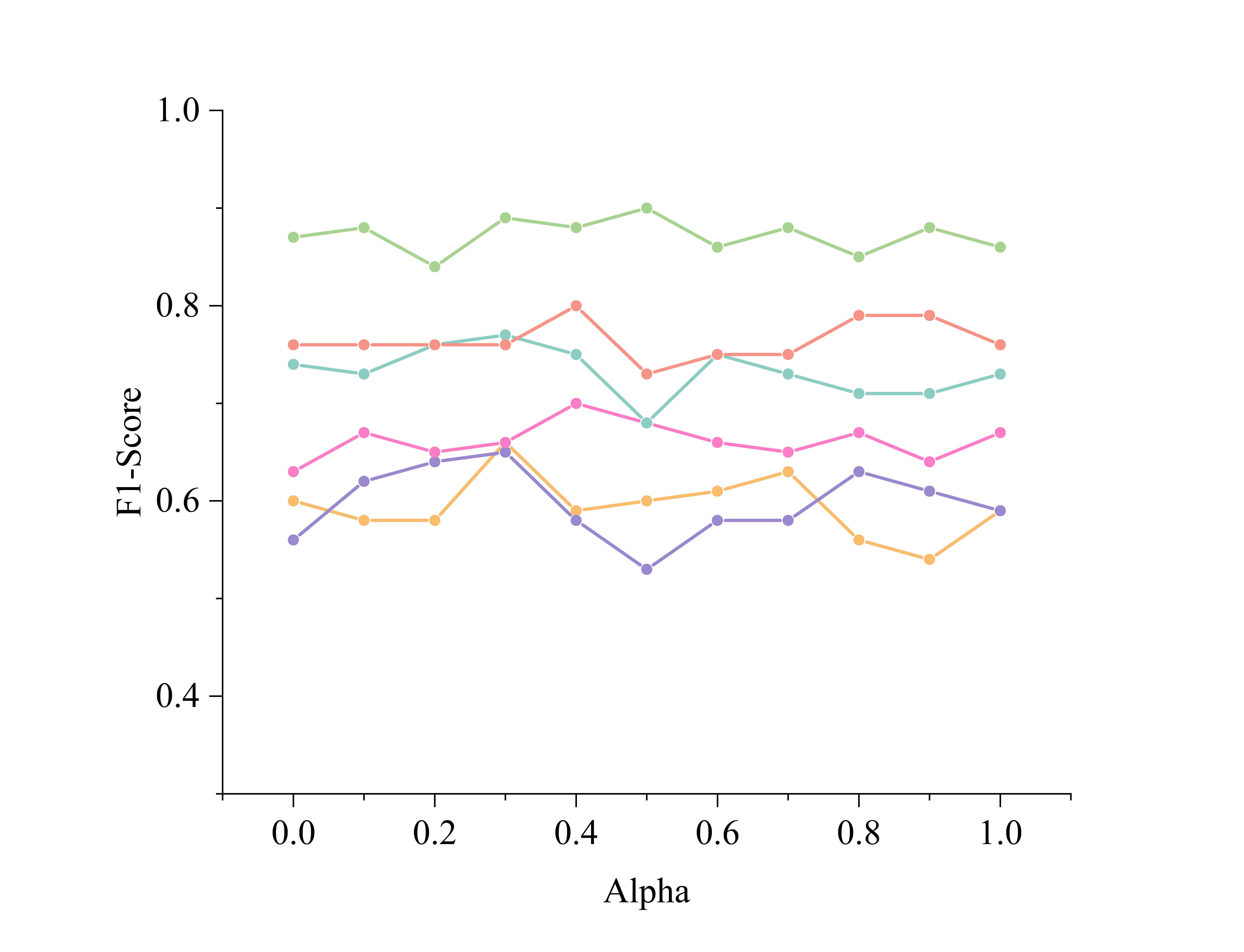}
        \caption{\textbf{PolitiFact}}
        \label{(a)}
    \end{subfigure}
    \hfill 
    \begin{subfigure}[b]{0.48\textwidth}
        \centering
        \includegraphics[width=\textwidth]{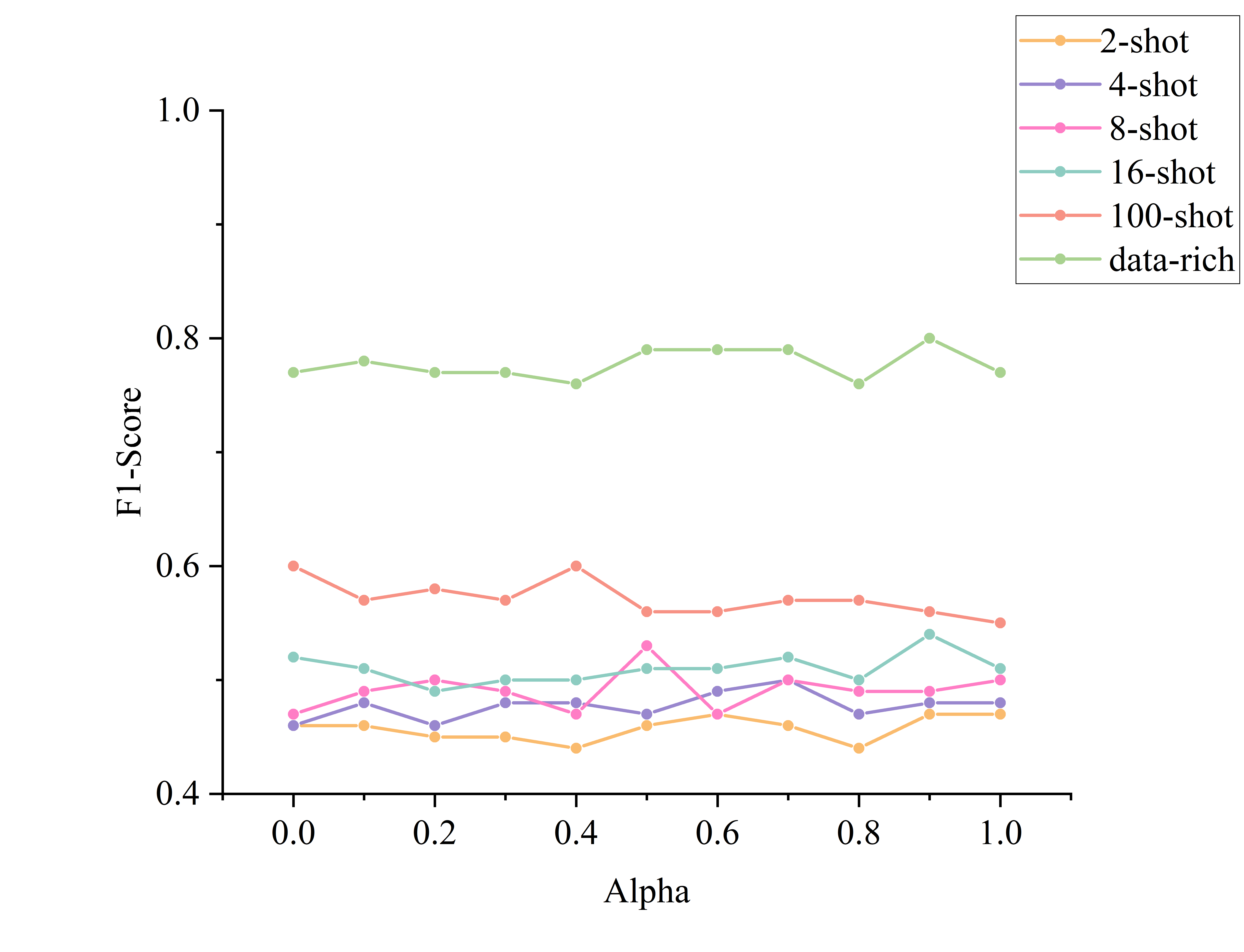}
        \caption{\textbf{GossipCop}}
        \label{(b)}
    \end{subfigure}
    \caption{Experimental results with different \( \alpha \) values.}
    \label{fig:3}
\end{figure}

\begin{figure}[H]
    \centering
    \includegraphics [width=7cm]{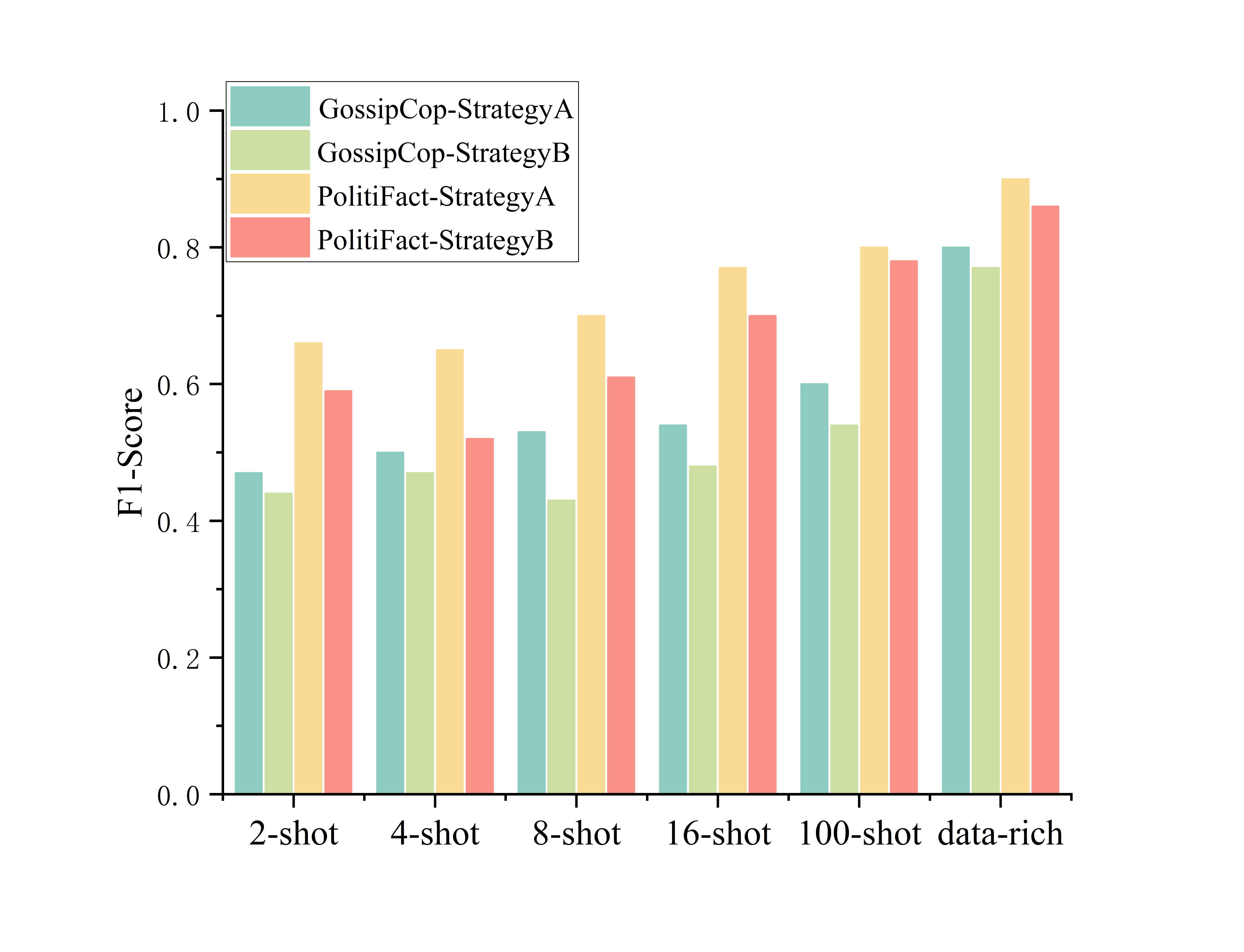}
    \caption{Comparison of F1 scores between modal fusion Strategy A and B.}
    \label{fig:4}
\end{figure}

\begin{itemize}
    \item \textbf{Strategy A: Adjusting the weight of MCA-computed versus non-MCA-computed features}
    \begin{itemize}
        \item We vary the contribution of these components by adjusting a weighting parameter, \( \alpha \), ranging from 0 to 1. \( \alpha = 0 \) utilizes only MCA-computed features, while \( \alpha = 1 \) uses only non-MCA-computed features.
        \item As shown in Figures \ref{(a)} and \ref{(b)}, the model performs best in both few-shot and data-rich settings when the contribution of non-MCA-computed components slightly exceeds that of MCA-computed components. This is likely because adjusting the weighting parameter \( \alpha \) can balance feature importance, and in few-shot settings, non-MCA-computed features provide regularization, reducing overfitting.

    \end{itemize}

    \item \textbf{Strategy B: Scaling both feature types equally}
    \begin{itemize}
        \item We utilize \( \alpha \) to scale both MCA-computed and non-MCA-computed features equally, ranging from 0 to 1.
        \item As shown in Figure \ref{fig:4}, Strategy A outperforms Strategy B, achieving an average F1 score 7.3\% higher across both datasets. This is likely because Strategy A selectively enhances the contribution of MCA-computed and non-MCA-computed features, optimizing feature usage, whereas Strategy B scales both feature types equally, potentially limiting its effectiveness.

    \end{itemize}
\end{itemize}
\vspace{-0.8cm}

\vspace{-1.3cm}

\subsection{Ablation Study}

In the ablation studies, we remove components from the AMPLE framework and retrain the model to assess their contribution. The results in Table \ref{tab:3} show that -EE significantly degrades performance. Emotional elements are crucial for capturing implicit emotional information, especially in FND where these emotions may be deceptive or inflammatory. Removing these elements limits the model's ability to leverage this information, leading to a drop in performance. -MCA also reduces performance by hindering the effective integration of multimodal features, resulting in information loss. -TM degrades performance more than -IM, indicating that text typically contains richer contextual information and better reflects the authenticity of news.

\vspace{-0.4cm}
\begin{table}[ht]
\caption{ "-EE" removes emotion elements, "-SA" removes automatic similarity adjustment, "-TM" removes text features not computed by MCA, "-IM" removes image features not computed by MCA, "-FM" removes fusion features computed by MCA, and "-RFM" removes fusion features not computed by MCA, using only text features from RoBERTa-base.}

\label{tab:3}
\centering
\setlength{\tabcolsep}{3pt}
\begin{tabular}{llllllll}
\bottomrule
  \multirow{2}{*}{\textbf{PolitiFact}} & \multicolumn{5}{c}{\textbf{few-shot(F1/Acc)}} &  \multirow{2}{*}{\textbf{data-rich} } \\\cline{2-6}
        &  \multicolumn{1}{c}{\textbf{2}} &  \multicolumn{1}{c}{\textbf{4}} &  \multicolumn{1}{c}{\textbf{8}} &  \multicolumn{1}{c}{\textbf{16}} & \multicolumn{1}{c}{\textbf{50}}  \\ 
\hline

  AMPLE & \textbf{0.66/0.67} & \textbf{0.65/0.67} & \textbf{0.70/0.71} & \textbf{0.77/0.78} & \textbf{0.80/0.80} & \textbf{0.90/0.90} \\
    -EE  & \underline{0.58}/\underline{0.60} & \underline{0.60}/\underline{0.60} & \underline{0.64}/\underline{0.66} & \underline{0.67}/\underline{0.68} & \underline{0.77}/0.78 & \underline{0.86}/\underline{0.87} \\
     -FM  & 0.51/0.54 & 0.46/0.50 & 0.63/0.63 & 0.66/0.66 & 0.73/0.74 & \underline{0.86}/\underline{0.87} \\
     -SA & 0.44/0.55 & 0.55/0.57 & 0.58/0.65 & 0.65/0.65 & 0.75/\underline{0.79} & 0.76/0.81 \\
     -TM & 0.43/0.53 & 0.51/0.57 & 0.55/0.63 & 0.60/0.61 & 0.73/0.77 & 0.75/0.78 \\
     -IM & 0.35/0.43 & 0.46/0.50 & 0.51/0.60 & 0.54/0.65 & 0.66/0.69 & 0.69/0.71 \\
     -RFM & 0.32/0.48 & 0.43/0.51 & 0.46/0.56 & 0.55/0.57 & 0.63/0.69 & 0.65/0.65 \\
\hline
\multirow{2}{*}{\textbf{GossipCop}}& \multicolumn{5}{c}{\textbf{few-shot(F1/Acc)}} &\multirow{2}{*}{\textbf{data-rich} }
\\\cline{2-6} 
 & \multicolumn{1}{c}{\textbf{2}}  & \multicolumn{1}{c}{\textbf{4}}  & \multicolumn{1}{c}{\textbf{8}} & \multicolumn{1}{c}{\textbf{16}} & \multicolumn{1}{c}{\textbf{100}} \\
\hline
 AMPLE & \textbf{0.47/0.55} & \textbf{0.50/0.57} & \textbf{0.53/0.64} & \textbf{0.54/0.65} & \textbf{0.60}/\underline{0.66} & \textbf{0.80/0.85} \\
     -EE  & 0.42/0.44 & \textbf{0.50}/\underline{0.54} & \underline{0.51}/\underline{0.59} & \underline{0.50}/0.53 & \underline{0.56}/0.61 & \underline{0.73}/\underline{0.76} \\
     -FM  & \underline{0.45}/0.46 & 0.44/0.47 & 0.39/0.40 & 0.47/0.51 & 0.55/0.58 & 0.68/0.71 \\
    -SA & 0.43/\underline{0.50} & \underline{0.45}/\textbf{0.57} & 0.50/0.51 & \underline{0.50}/\underline{0.59} & 0.55/\textbf{0.69} & 0.59/0.75 \\
 -TM & 0.39/0.43 & 0.43/0.50 & 0.48/0.55 & \underline{0.50}/0.55 & 0.53/0.65 & 0.53/0.70 \\
 -IM & 0.37/0.49 & 0.38/0.49 & 0.41/0.50 & 0.44/0.50 & 0.49/0.56 & 0.51/0.65 \\
 -RFM & 0.35/0.38 & 0.39/0.45 & 0.40/0.47 & 0.41/0.49 & 0.45/0.55 & 0.49/0.55 \\
\bottomrule

\end{tabular}
\end{table}
\vspace{-0.8cm}
\subsection{Emotion-based LLM study}
The LLM \cite{jiang2024large,ma2023large,wang2024instruction,xu2024team} offers a novel approach to calculating news emotion elements due to its powerful text comprehension capabilities. This study aims to leverage the advanced capabilities of LLMs to more accurately capture both explicit and implicit emotions, along with their complex semantics, and evaluates the performance of the ChatGPT-3.5 Turbo\footnote{\url{https://chat.openai.com}} model, developed by OpenAI, on the PolitiFact dataset. 



Given the large number of parameters in the ChatGPT-3.5 model and its non-open-source nature, we adopt the few-shot chain of thought prompting technique \cite{hu2024bad} (using two 4-shot configurations), which has been proven effective in downstream tasks. As shown in Table \ref{tab:5}, the emotion elements generated by LLM do not significantly improve model performance compared to the TextBlob method. This may be because LLM excels at text generation but is limited in capturing and integrating subtle emotions. Nonetheless, LLM shows potential in handling complex emotions and domain-specific feature fusion.
\vspace{-0.5cm}

\begin{table}[H]
\centering
\setlength{\tabcolsep}{7.5pt}
\caption{Inspired by \cite{hu2024bad}, we encode the text generated by ChatGPT-3.5 and concatenate it with \( T^c \) (CHAT\textsubscript{1}). We also use the MCA mechanism to merge the encoded text with \( T^c \) (CHAT\textsubscript{2}) and replace the emotion elements generated by TextBlob with those from ChatGPT-3.5, integrating them into the model (CHAT\textsubscript{3}).}\label{tab:5}
\centering
\renewcommand{\arraystretch}{1.5} 
\begin{tabular}{llllll}
\hline
  \multirow{2}{*}{\textbf{PolitiFact}} & \multicolumn{5}{c}{\textbf{few-shot(F1/Acc)} }   \\\cline{2-6}
& \multicolumn{1}{c}{\textbf{2 }} &  \multicolumn{1}{c}{\textbf{4}} &  \multicolumn{1}{c}{\textbf{8}} &  \multicolumn{1}{c}{\textbf{16}} & \multicolumn{1}{c}{\textbf{50}}  \\ 
\hline
 CHAT\textsubscript{1} & 0.58/0.59 & 0.62/0.63 & \underline{0.68}/\underline{0.68} & \underline{0.72}/\underline{0.73} & 0.78/0.78  \\
CHAT\textsubscript{2} & 0.58/0.59 & \underline{0.63}/\underline{0.65} & 0.67/0.67 & 0.71/0.72 & \underline{0.79}/\underline{0.79}  \\
CHAT\textsubscript{3} & \underline{0.62}/\underline{0.64} & 0.55/0.56 & 0.66/0.66 & 0.67/0.68 & 0.76/0.76  \\
AMPLE & \textbf{0.66}/\textbf{0.67} & \textbf{0.65}/\textbf{0.67} & \textbf{0.70}/\textbf{0.71} & \textbf{0.77}/\textbf{0.78} & \textbf{0.80}/\textbf{0.80}   \\
\hline

\end{tabular}
\end{table}

\section{Conclusion and Limitations}

This paper introduces the AMPLE framework, which leverages prompt learning in few-shot scenarios to alleviate the problem of data scarcity and enhances the authenticity analysis of fake news by incorporating emotional elements extracted by SAT. The designed MCA mechanism and the ability to perceive semantic similarity facilitate the integration and interaction of multimodal information. Extensive experiments based on benchmark datasets show that the AMPLE framework proposed in this paper exhibits obvious advantages in integrating emotional elements and multimodal fusion.
Although the improvement on some datasets is not significant, it still effectively enhances the accuracy of FND. This indicates that emotional elements hold potential in this task, particularly in datasets with specific emotional distributions. In addition, we explore that there is still much room for improvement in the prediction effect of combining the emotional elements generated by LLM with the small model.

However, this study has several limitations. First, concerning the combination of emotional elements and FND, alternative emotional feature extraction methods can be explored. Secondly, this study lacks analysis of other relevant information related to fake news, such as user information, comments, and retweets. Moreover, the potential impacts of LLM on the proposed method require further investigation. In future research, we aim to enhance the model's generalization capability by designing new feature extraction and fusion methods. Additionally, we plan to conduct more experimental studies on the application of LLM to further validate and improve our method.

\section*{Acknowledgements}
This work is funded by the Natural Science Foundation of Shandong Province under grant ZR2023QF151.


\bibliographystyle{splncs04}
\bibliography{mybibfile}

\end{document}